%% file: main.tex
\documentclass[runningheads]{llncs}
\usepackage[T1]{fontenc}
\usepackage{graphicx}
\usepackage{hyperref}
\usepackage{color}
\input{packages}
\begin{document}
\title{Evolutionary Bilevel Reward Shaping for Generalization in Reinforcement Learning}
\titlerunning{Evolutionary Bilevel Reward Shaping for Generalization in RL}
\author{Ekasit Usaratniwart
\inst{1}\orcidID{0009-0005-6981-6301} \and
Xilin Gao
\inst{1}\orcidID{0009-0005-8218-6375} \and
Marc Ong\inst{1}\orcidID{0009-0009-9667-3026} \and
Youhei Akimoto\inst{1,2}\orcidID{0000-0003-2760-8123}}
\authorrunning{E. Usaratniwart et al.}
%
\institute{
University of Tsukuba, Tsukuba, Japan
\and
RIKEN Center for Advanced Intelligence Project, Tokyo, Japan
\email{
    \{ekasit,xilin.gao,marc\}@bbo.cs.tsukuba.ac.jp\\
    akimoto@cs.tsukuba.ac.jp}
}
\maketitle              
\begingroup
\makeatletter
\renewcommand\thefootnote{}
\renewcommand\@makefntext[1]{\noindent #1}
\makeatother
\endgroup
\begin{abstract}
Reinforcement learning (RL) often suffers from performance degradation when deployed in environments that differ from those encountered during training. Existing techniques such as domain randomization (DR) mitigate this, but require access to diverse training environments and full trajectory observability, assumptions that fail in privacy-preserving or restricted scenarios where only scalar performance metrics are available. We propose Generalization via Evolutionary Reward Shaping (GERS), a bilevel optimization approach to improve generalization on unseen test environments using only scalar feedback from validation environments. At the lower level, an RL agent guided via a reward function shaped by the upper level learns a policy on a limited set of training environments with accessible trajectory data; at the upper level, CMA-ES optimizes the reward shaping parameters to maximize the cumulative unshaped reward on separate validation environments for which trajectory access is unavailable. Results on continuous control tasks indicate that GERS outperforms the standard RL baseline on unseen test environments. GERS performance is comparable to DR, despite DR treating the combined set of training and validation environments of GERS as a single training set that requires trajectory access, whereas GERS cannot access validation trajectories. These results confirm that GERS effectively enhances generalization under restricted data access constraints.
\end{abstract}

\section{Introduction}

Deep reinforcement learning has emerged as a powerful framework for optimizing decision-making in complex environments~\cite{sutton2018reinforcement,arulkumaran2017deep}. Despite its success in controlled simulations, a significant hurdle to its deployment in real-world problems remains the generalization gap~\cite{cobbe2019}. When a reinforcement learning (RL) agent interacts with a fixed or limited training environment, the learned policy inevitably overfits to that specific environment's dynamics~\cite{cobbe2019,zhang2018study}. Consequently, the policy often fails when deployed in test environments that differ even slightly from those used during training~\cite{risi2020procedural,packer2018assessing}.

Existing research on RL generalization often relies on Domain Randomization (DR)~\cite{tobin2017,akkaya2019solving} or Procedural Content Generation (PCG)~\cite{cobbe2020leveraging,risi2020procedural}, which typically require access to a wide variety of training environments. Other common approaches, such as Data augmentation~\cite{raileanu2021b} or specialized exploration policies~\cite{zisselman2023}, bypass the need for multiple environments; however, they are primarily designed to handle visual variations or complex exploration hurdles. In this paper, we step away from visual or exploration-based challenges to focus specifically on generalizing across shifting system dynamics. We tackle the problem where the fundamental mechanics of the environment change, meaning the test environment operates under a different transition probability, than the training environment. This specific focus is necessary for real-world applications where developers are prevented from observing diverse trajectories or interacting with multiple environments during training.

Our approach addresses a challenging scenario where standard generalization techniques are often inapplicable: trajectories from a limited set of training environments are fully observable, while trajectories from validation environments are unavailable. In this setting, the developer can optimize the policy using state-action trajectories observed in the training environments, but can only observe a single scalar performance metric (cumulative undiscounted reward) from the validation environments. Without access to the trajectories in the validation environments, the policy gradients can not be computed for these environments. Therefore, the scalar signals provided by the validation environments can be utilized only to steer the training indirectly.

A real-world example of this constraint is automated red teaming, where an RL policy is developed to identify network vulnerabilities~\cite{nguyen2023,hasegawa2024,schwartz2019autonomous}. In this application, the agent is trained in a local, transparent sandbox network (training environment) where every network hop and system log (as a state) is visible, allowing the developer to record and learn from full trajectories~\cite{standen2021cyborg}.
However, for security and data privacy reasons, the policy must be validated against a client's live production replica (validation environment), a black box where the specific sequence of states and actions must remain hidden to protect sensitive infrastructure~\cite{slagell2005sharing}. While the developer cannot access these private trajectories, they receive a discounted return in the form of a vulnerability score.
This constraint is common in decentralized or privacy-preserving settings where validation environments are available, yet distributed across different stakeholders who cannot share raw data but are willing to share performance outcomes.

We hypothesize that the generalization ability can be improved by utilizing a well-shaped reward function during the training, even if the training is performed on a limited number of training environments that have slightly different transition kernels from the test environments. Reward design is essential for high-performance RL, yet it is notoriously tedious to design reasonable reward functions manually~\cite{sutton2018reinforcement,ng1999}. Reward hacking, where a policy maximizes cumulative rewards through unexpected and undesired behavior, is a frequent consequence of suboptimal reward design~\cite{skalse2022defining}. By shaping the reward function to avoid such behaviors, the generalization ability of the trained policy can be significantly improved~\cite{langosco2022goal}. We conjecture that \emph{the cumulative unshaped rewards from validation environments help to shape a reward function automatically}. 

We investigate whether optimizing the reward function in the training environment can bridge this generalization gap using only scalar feedback from the validation environments.
For this purpose, we propose a bilevel RL approach combining proximal policy optimization (PPO)~\cite{schulman2017} for the lower-level policy optimization and covariance matrix adaptation evolution strategy (CMA-ES)~\cite{hansen2001,hansen2003,akimoto2020} for the upper-level reward-shaping parameter optimization. 
The upper-level CMA-ES maximizes the validation score obtained by the policy trained for the shaped-reward function, which repeatedly requires PPO training. To reduce the computational cost, we introduce parameter sharing between different lower-level PPO calls.

Empirical studies on Multi-Joint dynamics with Contact (MuJoCo) problems show that our approach enables the agent to achieve a high generalization performance in the unseen test environments (i.e., the generalization performance is as high as the training performance), whereas the baseline PPO exhibits overfitting to the training environments (a high performance on the training environments, but a significantly lower performance on the test environments). 
Moreover, we demonstrate that in a relaxed setting where the trajectories in the validation environments are available (i.e., the validation environments can be used as training environments, and DR approach is applicable), our approach without access to the trajectories on the validation environments performs competitively to DR approach using both training and validation environments as a large set of training environments, showing the effectiveness of the proposed approach.

\section{Problem Formulation}

Our problem is formulated as a contextual Markov decision process (CMDP). 
A CMDP is defined by the tuple
$\mathcal{M}_\omega=(\mathcal{S},\mathcal{A},T_\omega,\rho_\omega,r,\gamma),$
where $\mathcal{S}$ and $\mathcal{A}$ denote the state and action spaces,
respectively; $T_\omega:\mathcal S \times \mathcal A \times \mathcal S \to \R$ the
state transition probability distribution; $\rho_{\omega}:\mathcal S \to \R$ the initial
state distribution;
$r:\mathcal{S}\times\mathcal{A}\times\mathcal{S}\to\mathbb{R}$ the reward
function; and $\gamma\in[0,1)$ the discount factor.
Different from the standard Markov decision process (MDP), $T_\omega$ and $\rho_\omega$ are parameterized by a context parameter $\omega \in \Omega$. 

The agent interacts with CMDP environments as follows. In each episode, the context parameter $\omega \in \Omega$ is selected. The way it is selected depends on situations and approaches. For a given CMDP environment $\mathcal{M}_\omega$, the agent begins in an initial state $s_0 \sim \rho_\omega$ and at each timestep $t$, executes an action $a_t \in \mathcal{A}$ that yields a reward $r_t= r(s_t,a_t)$ and the next state $s_{t+1} \sim T_\omega(\cdot \mid s_t,a_t)$. A policy $\pi$, defined as the conditional probability distribution $\pi(a \mid s)=\Pr(A_t=a \mid S_t=s)$, determines the agent's action at each state. 
In the standard MDP setting, a RL agent typically learns an optimal policy $\pi^\star$ that maximizes the expected discounted return,
\begin{equation}
    \pi^{\star} = \argmax_\pi R(\pi; r, \omega) := \E_{\pi,\mathcal{M}_\omega} \left[ \sum_{t=0}^\infty \gamma^t r (s_t,a_t)\right],
\end{equation}
where the discounted return $R$ depends on the context parameter $\omega$ and the reward function $r$.

In this study, we address the generalization problem in a contextual MDP setting. The training agent has access to a finite set of environments, $\mathbb{M}_\text{train} = \{\mathcal{M}_{\omega_1}, \dots, \mathcal{M}_{\omega_{N_\text{train}}}\}$, called \emph{training environments}. The agent optimizes its policy on these $N_\text{train}$ environments. For example, domain randomization~\cite{raileanu2021b,tobin2017} trains the policy by sampling a training environment uniform-randomly from $\mathcal{M}_{\omega_{N_\text{train}}}$ at each episode. It corresponds to maximizing the discounted return averaged on $\mathcal{M}_{\omega_{N_\text{train}}}$, namely,
$J_\mathrm{train}(\pi) = \frac{1}{N_\text{train}} \sum_{i=1}^{N_\text{train}} R(\pi; r, \omega_i).$
During the training, the agent has full access to the trajectory $(s_0,\allowbreak a_0,\allowbreak r_0,\allowbreak s_1,\allowbreak a_1,\allowbreak r_1,\allowbreak \dots,\allowbreak s_{T-1},\allowbreak a_{T-1},\allowbreak r_{T-1},\allowbreak s_{T})$ of each episode, where $T$ denotes the number of interaction steps performed before termination.  Our objective, however, is not to maximize the discounted returns on these training environments, but to maximize the cumulative undiscounted rewards in unseen test environments.  Given the distribution of the context parameter, $\omega \sim \mathcal{P}_\text{test}$, the generalization ability is often measured by the cumulative undiscounted return expected over $\omega \sim \mathcal{P}_\text{test}$, namely,  $ J_\text{gen}(\pi) = \E_{\omega \sim \mathcal{P}_\text{test}}\left[\sum_{t=0}^\infty r(s_t,a_t; \omega) \right].$ Even if $\omega_1, \dots, \omega_{N_\text{train}} \sim \mathcal{P}_\text{test}$, due to the limited number $N_\mathrm{train}$ of training environments, maximizing $J_\mathrm{train}$ typically results in an overfitted policy, and we observe a generalization gap, $J_\mathrm{train}(\pi) \gg J_\mathrm{gen}(\pi)$.

This study addresses the RL generalization problem by utilizing partial access to additional validation environments, $\mathbb{M}_\text{val} = \{\mathcal{M}_{\omega^v_{1}}, \dots, \mathcal{M}_{\omega^v_{N_{\mathrm{val}}}}\}$, where $N_{\mathrm{val}} \gg N_{\mathrm{train}}$. 
We suppose that the agent can query the cumulative return of its policy, on each $\mathcal{M}_{\omega^v_{i}} \in \mathbb{M}_\text{val}$. 
However, we cannot observe a trajectory on these environments.
Therefore, the output of the validation environments cannot be used for standard policy gradient approaches. 
Instead, we utilize the average of these validation scores $ J_\mathrm{val}(\pi) = \frac{1}{N_\text{val}} \sum_{i=1}^{N_\text{val}} \left[\sum_{t=0}^T r (s_t,a_t; \omega^v_i)\right] $
to optimize the \emph{reward function for the training environment}.\footnote{Note that both $J_\mathrm{train}$ and $J_\mathrm{val}$ can be seen as Monte-Carlo estimates of $J_\mathrm{gen}$ if $\mathbb{M}_\mathrm{train}$ and $\mathbb{M}_\mathrm{val}$ are sampled independently from $\mathcal{P}_\text{test}$. However, because $N_\mathrm{train}$ and $N_\mathrm{val}$ are limited, they can differ significantly.}

\section{Related Work}
\subsubsection{Generalization in RL}

Existing approaches to generalization in RL can be categorized into three classes~\cite{kirk2023}, The first class includes data augmentation~\cite{raileanu2021b}, domain randomization (DR)~\cite{tobin2017}, and unsupervised environment design (UED)~\cite{dennis2021emergentcomplexityzeroshottransfer}, which mainly work by diversifying or adapting the training environment. 
The second class encompasses regularization, which constrains policy learning to prevent overfitting, and learning invariances, which guide the policy to capture features that remain consistent across environments~\cite{cobbe2019,ghosh2021}. 
The third class addresses RL-specific challenges, utilizing architectural changes, such as decoupling the policy and value functions~\cite{cobbe2021phasic,raileanu2021a}.

While these methods have meaningfully pushed forward the field of RL generalization, most are not well-suited for our specific problem: generalizing to unseen transition dynamics when environment access is limited. Techniques such as data augmentation~\cite{raileanu2021b}, regularization~\cite{cobbe2019}, and architectural modifications~\cite{cobbe2021phasic,raileanu2021a} are largely designed to handle visual variation, encourage exploration, or reduce overfitting to a fixed state space. As a result, they do not directly tackle the core challenge we face, namely, differences in the underlying transition probability between training and test environments.

Methods that do alter environment dynamics, such as DR and UED, align more closely with our objective but rely on assumptions that violate our constraints. DR trains agents across a wide distribution of randomized parameters to encourage robust policies, while UED actively generates an automated curriculum of environments tailored to the agent's current capabilities. However, both DR and UED critically assume that the developer has access to a fully parameterized simulator and the freedom to continuously generate, alter, and access a vast diversity of training scenarios. In contrast, our work addresses a more constrained setting: the agent learns from a limited, fixed set of training environments and must adapt to unseen dynamics relying solely on episodic performance feedback from validation environments, entirely without access to their  trajectories or the ability to freely generate new environments.

\subsubsection{Reward Shaping}

Reward shaping is traditionally employed to improve RL sample efficiency and mitigate reward hacking~\cite{fu2025}, typically by accelerating convergence while preserving the original optimal policy~\cite{ng1999,devidze2022}. Recently, this concept has been extended to bilevel reward optimization frameworks. For example, in reinforcement learning from human feedback (RLHF)~\cite{christiano2017deep}, a reward model is optimized to align the agent's behavior with human preferences. Similarly, meta-learning approaches such as meta-gradient RL~\cite{xu2018meta} and bilevel intrinsic reward learning~\cite{stadie2020} dynamically adjust reward parameters to maximize returns or enhance exploration efficiency.

However, the fundamental objectives of these existing bilevel approaches are restricted to either aligning with expert supervision or maximizing performance strictly within the training distribution. In contrast, while we adopt the methodology of bilevel reward shaping, our objective is fundamentally different: we optimize the reward function specifically to guide the agent away from overfitting, thereby enhancing its generalization performance across unseen environments.

\section{Proposed Approach}
We propose a novel approach to improve the generalization gap by solving a bilevel reward shaping problem utilizing the validation score $J_\text{val}(\pi)$.

\subsection{Reward Shaping as Bilevel RL}
Our main idea to improve the generalization performance is to shape the reward function and use it to optimize the policy on the training environments. Let $\hat{r}_{\nu}: \mathcal{S} \times \mathcal{A} \to \R$ be a parametric model of a reward function, 
where $\mathcal{S}$ and $\mathcal{A}$ denote the continuous state and action spaces of the environment, respectively. The reward shaping parameter vector is denoted as $\nu \in \mathbb{R}^{d_\nu}$, where $d_\nu$ represents the dimensionality of the feature vector space. During the training, instead of using the original reward function $r_{\omega_i}$ in $\mathcal{M}_{\omega_i} \in \mathbb{M}_\text{train}$, the agent relies on the shaped reward function $\hat{r}_{\nu}$. That is, the discounted return on $\mathcal{M}_{\omega_i}$ is replaced with $R(\pi; \hat{r}_{\nu}, \omega_i)$. The objective function for the policy training is then 
$\hat{J}_\text{train}(\pi; \nu) = \frac{1}{N_\text{train}} \sum_{i=1}^{N_\text{train}} R(\pi; \hat{r}_{\nu}, \omega_i). $
The optimal policy $\pi^{\nu\dagger} = \argmax_{\pi} \hat{J}_\text{train}(\pi; \nu)$ under this objective depends on the reward-shaping parameter $\nu$. Therefore, we optimize $\nu$ so that the optimal policy $\pi^{\nu\dagger}$ under the shaped reward function $\hat{r}_{\nu}$ maximizes the generalization performance $J_\text{gen}(\pi)$. 

This reward optimization problem is formulated as a bilevel RL problem. 
The lower-level problem optimizes the policy, which is modeled by a neural network $\pi_\theta$ with a policy parameter $\theta \in \mathbb{R}^{d_\theta}$, where $d_\theta$ represents the total number of optimized weights and biases in the lower-level policy network, under the shaped reward function $\hat{r}_\nu$ determined by the upper-level problem. 
The upper-level problem optimizes the shaped reward function $\hat{r}_\nu$ so that the optimal policy $\pi^{\nu\dagger}$ maximizes the generalization performance $J_\text{gen}(\pi)$. Because $J_\text{gen}(\pi)$ is unavailable, we utilize $J_\text{val}(\pi)$ instead. We hope that the policy that performs well on validation levels, despite that it is not trained on them, will generalize well on unseen test environments.
To this end, our bilevel reward shaping problem is formulated as a bilevel RL, \footnote{Note that the shaped reward is used only in the training environments.}

\begin{equation}
\begin{aligned}
      \max_\nu \quad & \Jvl(\pi_{\theta^{\star}(\nu)}) \\
      \st \quad&\theta^{\star}(\nu) = \argmax_\theta \Jtrh(\pi_{\theta}; \nu).
\end{aligned}
\end{equation}

\begin{algorithm}[t]
\caption{Evolutionary Reward Shaping}\label{alg:rewardshape}
\begin{algorithmic}[1]\small
\State Perform $K_\mathrm{init}$ steps of PPO on training environment $\mathcal{M}_\text{train}$ with original reward $r$ and obtain optimized policy parameter $\theta^{(1)}$
\For{$g = 1, \dots, G$}
    \State {Sample candidates $\{\nu_i^{(g)}\}_{i=1}^\lambda$ from CMA-ES}
    \For{\textbf{each} $i = 1, \ldots, \lambda$ \textbf{in parallel}}
        \State Initialize the PPO agent with $\theta^{(g)}$ (both Actor and Critic)
        \State\parbox[t]{\dimexpr\linewidth-\algorithmicindent-\algorithmicindent}{Perform $K$ steps of PPO on $\mathbb{M}_\text{train}$ with $\hat r_{\nu_i^{(g)}}$}
        \State\parbox[t]{\dimexpr\linewidth-\algorithmicindent-\algorithmicindent}{Evaluate $\E_{\pi_{\theta_i^{(g)}}} \left[ \sum_{t=0}^T r(s_t, a_t;\omega) \right]$ of the trained $\theta_i^{(g)}$ on $\mathbb{M}_{\text{val}}$}
        \State  Compute the validation score $f(\nu_i^{(g)}) = \Jvl(\pi_{\theta_{i}^{(g)}})$
    \EndFor
    \State Update CMA-ES with $\{(\nu_i, f(\nu_i^{(g)})\}_{i=1}^{\lambda}$
    \State \parbox[t]{\dimexpr\linewidth-\algorithmicindent}{$i^\star \gets \argmax_{i \in \{1,\ldots,\lambda\}} f(\nu_i^{(g)})$; $\theta^{(g+1)} \gets \theta_{i^\star}^{(g)}$}
\EndFor
\Ensure Optimized policy $\pi_{\theta^{(G+1)}} = \pi_{\theta_{i^\star}^{(G)}}$, optimized reward $\hat r_{\nu_{i^\star}^{(G)}}$
\end{algorithmic}
\end{algorithm}
\subsection{\ppnamefull}
We proposed \ppnamefull, which integrates the covariance matrix adaptation evolution strategy (CMA-ES)~\cite{hansen2001,hansen2003,akimoto2020} for the upper-level reward-shaping optimization and proximal policy optimization (PPO)~\cite{schulman2017} for the lower-level policy optimization . 

\Cref{alg:rewardshape} presents the pseudocode of the proposed approach.
From the upper-level perspective, the variable to be optimized is the reward shaping parameter $\nu$, and the objective function is a map $f: \nu \mapsto \Jvl(\pi_{\theta^{\star}(\nu)})$. Each objective function evaluation requires solving the lower-level RL problem with a shaped reward function $\hat{r}_{\nu}$ to approximate $\theta^{\star}(\nu)$. For the moment, we suppose that $f(\nu)$ can be approximated for each $\nu$, for which the detail is explained later. Then, all we need is to simply apply the CMA-ES to optimize $\nu$ with $f$ as the objective function. 
The reason for choosing an evolutionary approach (CMA-ES) as the upper-level solver is twofold.
First, we cannot observe trajectories in the validation environments, which prevents us from applying approaches based on hyper policy gradients~\cite{kudo2026icaps,thoma2024,gupta2023behavior,stadie2020}. On the other hand, evolutionary approaches typically require only the objective function values, which perfectly match our problem setting.
Second, the computational bottleneck of the bilevel approach is the number of upper-level objective function evaluations. Each upper-level objective function evaluation requires solving a lower-level RL task. Though we incorporate techniques to accelerate the lower-level RL, parallel evaluation is almost necessary for practical use. Evolutionary approaches are parallel-evaluation-friendly. In the case of the CMA-ES that we employed in this paper, it has been reported that the number of iterations to reach the same target performance decreases as the number of candidate solutions, $\lambda$, increases~\cite{akimoto2020}. Therefore, by setting $\lambda$ to be the maximum number of parallel objective evaluations, we gain speed-up in wall clock time.

The upper-level objective function evaluation is costly if we perform the lower-level RL from scratch each time. 
To reduce the computational time for the lower-level RL, we use a pretrained policy as the initial policy for each lower-level RL run.\footnote{
More precisely, we inherit both the Actor and Critic networks of a PPO agent. 
In PPO, the Actor network models the policy and the Critic network models the value function that is used to compute the policy gradient. As the Critic network is essential for the training of the Actor network, it is insufficient to maintain only the Actor network to accelerate the training. Because the Actor and Critic networks are coupled, we inherit the parameters for both networks together. Hereafter, with a slight abuse of notation, the policy parameter $\theta$ represents both the parameters for the Actor and Critic networks. } 
Before the CMA-ES run, we first perform the pretraining of the policy in the training environments with the original reward functions. 
The obtained policy $\pi_{\theta^{(1)}}$ is used as the initial policy for $\lambda$ PPO runs in the first iteration of the CMA-ES. Each PPO uses a shaped reward function generated by the CMA-ES that is different from the original reward function. Therefore, the optimal policy parameter for each PPO run generally deviates from $\theta^{(1)}$. However, if the shaped-reward functions are close to the original reward function, we expect that $\theta^{(1)}$ plays as a reasonable initial policy parameter because $\Jtrh$ is recognized as an estimate of $J_\mathrm{test}$. 
After parallel PPO runs, we obtain the policy $\pi_{\theta_{i}^{(1)}}$ optimized for the shaped reward $\hat{r}_{\nu_i^{(1)}}$ for $i = 1, \dots, \lambda$. 
They are evaluated in the validation environments, and the policy with the best validation score among $\lambda$ current policies is selected as the next initial policy $\pi_{\theta^{(2)}}$. 
We expect that $\pi_{\theta^{(g)}}$ tends to converge as the CMA-ES itself converges because the shaped-reward functions generated by the CMA-ES eventually become similar, and so do the optimal policies for these shaped reward functions. 
We remark that the reason for not using the best-so-far policy as the initial policy is to prevent from using a policy that achieved a high validation performance by chance due to the stochasticity of the environments as well as the policy.

\section{Experiments}
We investigate whether GERS improves the agent's generalization ability beyond the training environments. 
Specifically, our goal is to elucidate whether GERS leads to better performance on unseen environments compared to the baseline.
To address this question, we examine whether GERS achieves higher performance than the baseline when tested on unseen environments, which directly reflects its generalization ability. 

\subsection{Task Description}
We evaluate~\ppabbr~on four MuJoCo continuous control environments (v5)~\cite{towers2025gymnasium}: \textit{InvertedPendulum}, \textit{Hopper}, \textit{HalfCheetah}, and \textit{Ant}. 
In all tasks, RL agents learn policies within continuous state and action spaces to maximize episodic return. 
To optimize the reward shaping parameters, we construct validation environments $\mathbb{M}_\text{val}$ by simultaneously scaling two physical parameters, indicated in~\Cref{tab:physical_params}, using a coefficient multiplier $c \in [0.1, 3.0]$. The agents are trained exclusively on a single training environment $\mathbb{M}_\text{train} = \{\mathcal{M}_{\omega_1}\}$, corresponding to the default setting  with physical parameter multiplier $c = 1.0$ applied to both parameters. Throughout all experiments, the training, and validation sets are strictly separated. Validation environments remain fixed during the upper-level optimization to ensure consistent performance estimation, and results are reported as average returns over multiple episodes to reduce stochastic variance.

\subsection{Experimental Settings}

\subsubsection{Pretraining Stage:} 
We pretrain a PPO agent for $K_{\mathrm{init}} = \num{3e7}$ timesteps. The resulting policy serves two roles: a baseline for comparison, and the initial policy for~\ppabbr.  

\subsubsection{Reward Shaping Optimization (\ppabbr~Stage):}
 The shaped reward function evaluated during the lower-level reinforcement learning phase is linearly parameterized as follows:
\begin{equation}
\hat{r}_{\nu^{(t)}}(s,a) = r(s,a) + \inner{\nu^{(t)}}{\pphi(s,a)},
\label{eq:reward-shaping-formular}
\end{equation}
where $\pphi(s,a) = [s^\top, a^\top]^\top \in \mathbb{R}^{d_s + d_a}$ is a composite feature vector constructed by concatenating the state observation $s \in \mathcal{S}$ and the action vector $a \in \mathcal{A}$, and $\inner{\cdot}{\cdot}$ denotes the standard inner product.
To maintain numerical stability and ensure comparable feature scaling, each element of the feature vector is normalized (see \Cref{sec:normalization}) before computing the weighted sum.

The fitness of each CMA-ES candidate is computed as the mean episodic return over a set of 25 validation environments. The validation environments are generated by uniformly sampling physical parameters, using a coefficient multiplier $c \in [0.1, 3.0]$. Each validation environment is evaluated for 4 episodes, resulting in a total of 100 validation episodes per candidate.
The 25 physical parameter variations for validation environments are generated once and remain strictly constant throughout the entire CMA-ES process. 

We resample the random seeds\footnote{In MuJoCo environments, the seed specifically determines the initial state position for the episode.} at the start of every new generation. Within any single generation, these seeds (total 100 seeds, one per levels) are held identical across all candidates to ensure their fitness scores are directly comparable and to avoid overfitting to specific initial conditions.\footnote{If the same seed were reused across generations, the agent would repeatedly encounter identical initial states, potentially leading to overfitting to those specific trajectories.}

In this experiment, CMA-ES is executed for $G = 100$ iterations with initial distribution $\mathcal{N}(\mm , 0.1^2 \II)$ with $\mm \sim \mathcal{N}(\bm 0, \II)$ and population size $\lambda = 60$. The reward parameter is constrained to $[-5, 5]^{d_{\nu}}$ and this box constraint is handled by a mirroring \cite{Yamaguchi2018gecco}.
 We run each lower-level PPO training process for $K=81920$ steps using the shaped reward for each candidate solution in parallel. We use the same training environment, $\M_{\mathrm{train}} = \{\mathcal{M}_{\omega_1}\}$ for both the pretraining stage and the CMA-ES stage.

 The proposed method uses CMA-ES, specifically the variant dd-CMA-ES ~\cite{akimoto2020}, to optimize the reward shaping parameter vector $\nu \in \mathbb{R}^{d_\nu}$, where $d_\nu = d_s + d_a$, with $d_s$ and $d_a$ denoting the dimensions of the continuous state vector and action space vector, respectively.

\subsubsection{Testing and Comparison:}
In practice, generalization is estimated using unseen test environments 
$\mathbb{M}_\mathrm{test}=\{\mathcal{M}_{\omega_i^t}\}_{i=1}^{N_\mathrm{test}}$, 
sampled from $\mathcal{P}_\mathrm{test}$, independently from the training set $\mathbb{M}_\mathrm{train}$ and validation set 
$\mathbb{M}_\mathrm{val}$.
The test score is 
$J_\mathrm{test}(\pi)=\frac{1}{N_\mathrm{test}}\sum_{i=1}^{N_\mathrm{test}} \left[\sum_{t=0}^T r(s_t, a_t; \omega)\right]$. 
We generate $N_\mathrm{test}=100$ environments by sampling the multiplier 
$c\in[0.1,3.0]$, which remain unseen during training. Each method is evaluated 
using a deterministic policy for 4 episodes per environment (400 episodes total), 
and the average return is reported. The same test set and protocol are used for both our method and PPO for fair comparison.

\subsubsection{Domain Randomization Setting:} 
\begin{figure*}[t]
\centering

\begin{subfigure}[t]{0.5\textwidth}
    \centering
    \includegraphics[width=\linewidth]{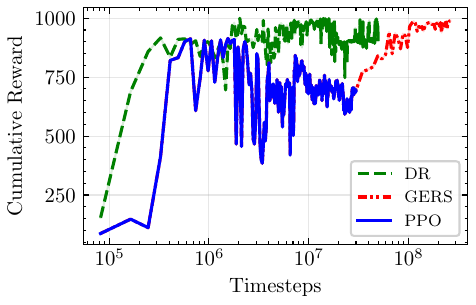}
    \caption{InvertedPendulum}%
\end{subfigure}%
\begin{subfigure}[t]{0.5\textwidth}
    \centering
    \includegraphics[width=\linewidth]{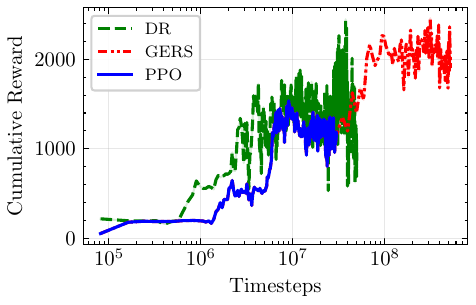}
    \caption{Hopper}%
\end{subfigure}%
\\
\begin{subfigure}[t]{0.5\textwidth}
    \centering
    \includegraphics[width=\linewidth]{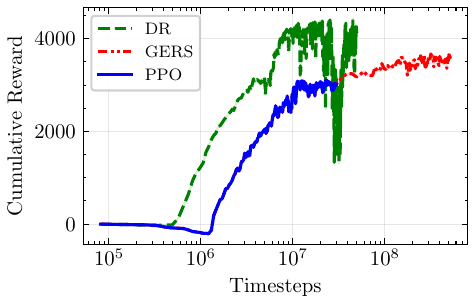}
    \caption{HalfCheetah}%
\end{subfigure}%
\begin{subfigure}[t]{0.5\textwidth}
    \centering
    \includegraphics[width=\linewidth]{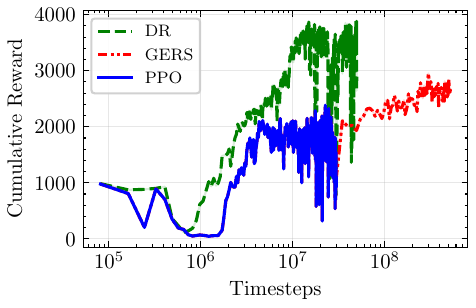}%
    \caption{Ant}%
\end{subfigure}

\caption{Test performance comparison of PPO baseline, DR, and GERS on MuJoCo tasks. 
Each subplot reports the mean test return on unseen environments with modified physical parameters. 
PPO is trained on a single fixed environment, DR uses randomized environments with full trajectory access, and GERS relies only on scalar validation returns. 
GERS achieves comparable or improved generalization under restricted information access.}
\label{fig:all_results}
\end{figure*}

We further benchmark our proposed method against a DR baseline optimized via PPO with $\num{5e7}$ timesteps \footnote{We choose this specific number because from~\Cref{fig:all_results} observed that the DR test performance becomes completely stable by this point; training it for any additional timesteps does not improve its results.}. To construct the training set for DR, we utilize the 25 levels previously reserved for validation plus a training level $\mathbb{M}_\mathrm{train} \cup \mathbb{M}_\mathrm{val}$, totaling 26 levels. Notably, DR is incompatible with our experimental criteria, as they demand full trajectory access across all environments. In contrast, our setting strictly restricts validation environments to providing only a single scalar return (the episodic reward).

\begin{figure*}[t]
\centering
\begin{subfigure}{0.5\hsize}
\includegraphics[width=\hsize]{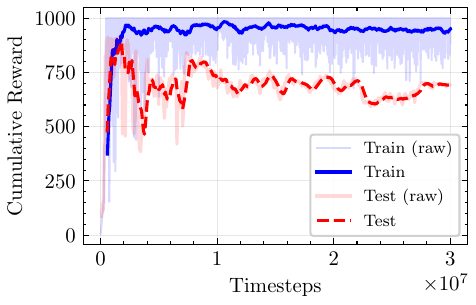}
\caption{InvertedPendulum}%
\end{subfigure}%
\begin{subfigure}{0.5\hsize}
\includegraphics[width=\hsize]{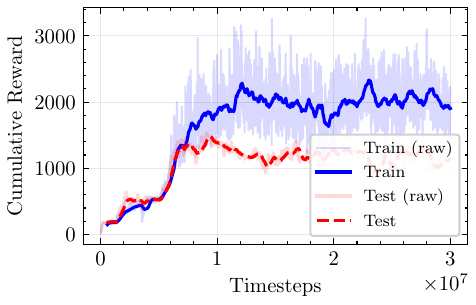}
\caption{Hopper}%
\end{subfigure}%

\begin{subfigure}{0.5\hsize}
\includegraphics[width=\hsize]{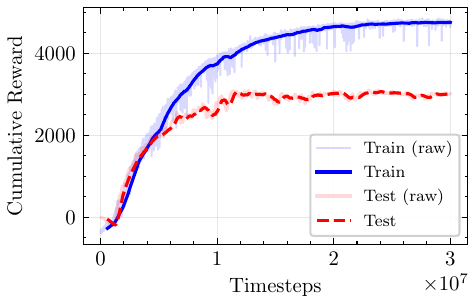}
\caption{HalfCheetah}%
\end{subfigure}%
\begin{subfigure}{0.5\hsize}
\includegraphics[width=\hsize]{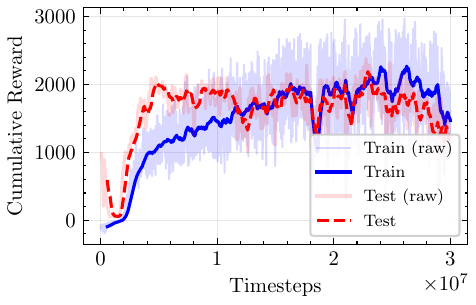}
\caption{Ant}%
\end{subfigure}%
\caption{Overfitting behavior of PPO trained on a single fixed environment. 
Training performance increases steadily, while test performance on unseen environments remains lower, indicating a clear generalization gap.}
\label{fig:ppo_overfitting}

\end{figure*}

\begin{table*}[t]
\centering
\caption{Final test performance across 5 trials per task. Values are mean $\pm$ std.}
\label{tab:final_test_mean_std}
\begin{tabular*}{\linewidth}{@{\extracolsep{\fill}} l l l l}
\toprule
Task & BasePPO & DR & GERS \\
\midrule
InvertedPendulum
& $\phantom{0}527.29 \pm \phantom{00}82.72$
& $\phantom{0}906.99 \pm \phantom{00}39.02$
& $\phantom{0}926.51 \pm \phantom{00}49.47$ \\
Hopper
& $\phantom{0}621.95 \pm \phantom{0}125.20$
& $1260.41 \pm \phantom{0}298.49$
& $1673.58 \pm \phantom{0}177.01$ \\
HalfCheetah
& $2563.20 \pm 1462.29$
& $4956.37 \pm 1758.56$
& $3992.22 \pm \phantom{0}618.89$ \\
Ant
& $2154.64 \pm \phantom{0}456.25$
& $2740.69 \pm \phantom{0}511.07$
& $2565.73 \pm \phantom{0}372.75$ \\
\bottomrule
\end{tabular*}
\end{table*}

\subsection{Results and Discussion}
\subsubsection{Baseline PPO fails to generalize:} 
\Cref{fig:ppo_overfitting} illustrates the performance of the baseline PPO trained for $\num{3e7}$ steps on a single training level. Although PPO achieves stable performance on the training environment, its test rewards remain consistently low across unseen levels. This indicates that PPO tends to overfit to the training environment and struggles to generalize to new unseen environments.\footnote{Although, in principle, the PPO baseline could be trained for the same total number of timesteps as the proposed method, in practice this would be time-consuming. The empirical observation (\Cref{fig:ppo_overfitting}) shows that the PPO policy maintains a stable performance plateau at $\num{3e7}$ steps, indicating near-optimal convergence under the original reward. Further training beyond this point does not yield noticeable improvement. Therefore, we consider the $\num{3e7}$ step PPO a reasonable and fair baseline for comparison.}

\subsubsection{GERS Improves Generalization Performance}
\Cref{fig:all_results} compares the test performance of GERS and the baseline PPO on four MuJoCo tasks. Across all tasks, the baseline PPO (blue) improves in the early stage but later becomes unstable or stops improving on the test environments, which is consistent with overfitting to the fixed training setting. In contrast, after the CMA-ES stage starts, GERS (red) shows a clear and consistent increase in test return. 
In InvertedPendulum, GERS reaches near-maximum test performance and remains stable.\footnote{For Inv.~Pendulum, the CMA-ES optimization was terminated early before reaching the maximum number of generations ($T=100$), as CMA-ES detects stagnation criterion, when the recent best fitness fails to exceed the historical median fitness over a generation.} In Hopper, HalfCheetah, and Ant, GERS exceeds the baseline and reaches higher test returns. These results suggest that the reward shaping optimized using validation returns helps the policy learn behaviors that transfer better to unseen environments with modified physical parameters (mass, friction, or gravity) indicated in \Cref{tab:physical_params}, leading to improved generalization.

\Cref{tab:final_test_mean_std} presents the final test performance of PPO, DR, and GERS across four continuous control tasks over five independent trials. The results demonstrate that both DR and GERS consistently improve generalization compared to the BasePPO baseline, which yields the lowest mean returns across all environments. Specifically, GERS achieves the highest mean returns on the InvertedPendulum and Hopper tasks, while DR performs best on the more complex HalfCheetah and Ant environments. Further details and extended analysis regarding these performance metrics are provided in the \Cref{sec:stats_analyse}

\subsubsection{Domain Randomization:}
The results, shown in \Cref{fig:all_results} and \Cref{tab:final_test_mean_std}, indicate that DR achieves mixed test performance compared to \ppabbr, outperforming on HalfCheetah and Ant, but underperforming on InvertedPendulum and Hopper. The two approaches, however, differ fundamentally in how the additional environments are used. In DR, the same set of environments that serve as validation levels in \ppabbr~are incorporated directly into the training set, and full state–action trajectories from these levels are used for policy optimization. In contrast, \ppabbr{} treats these exact same levels as black-box validation environments (no trajectories are accessed), and only scalar episodic returns are observed to guide the upper-level reward optimization. Therefore, achieving comparable performance demonstrates that the proposed approach attains similar robustness to DR while operating under restrictive information access.

\subsubsection{In-Distribution and Out-Of-Distribution Generalization Performance:}

\begin{figure*}[!t]
\centering

\includegraphics[width=0.88\textwidth]{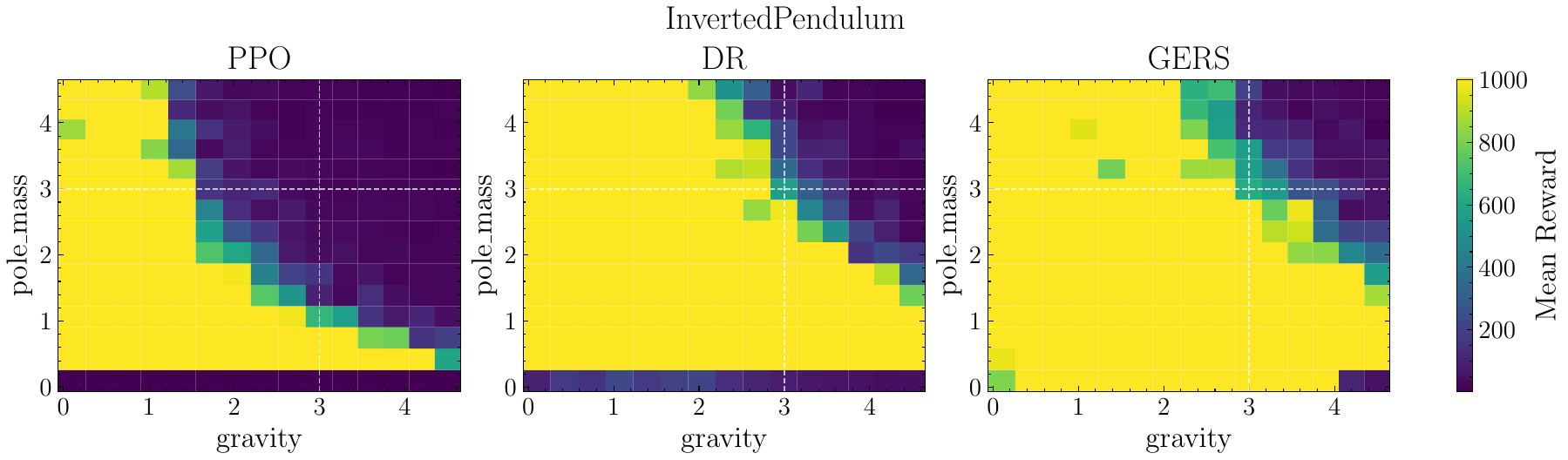}

\vspace{-0.15em}

\includegraphics[width=0.88\textwidth]{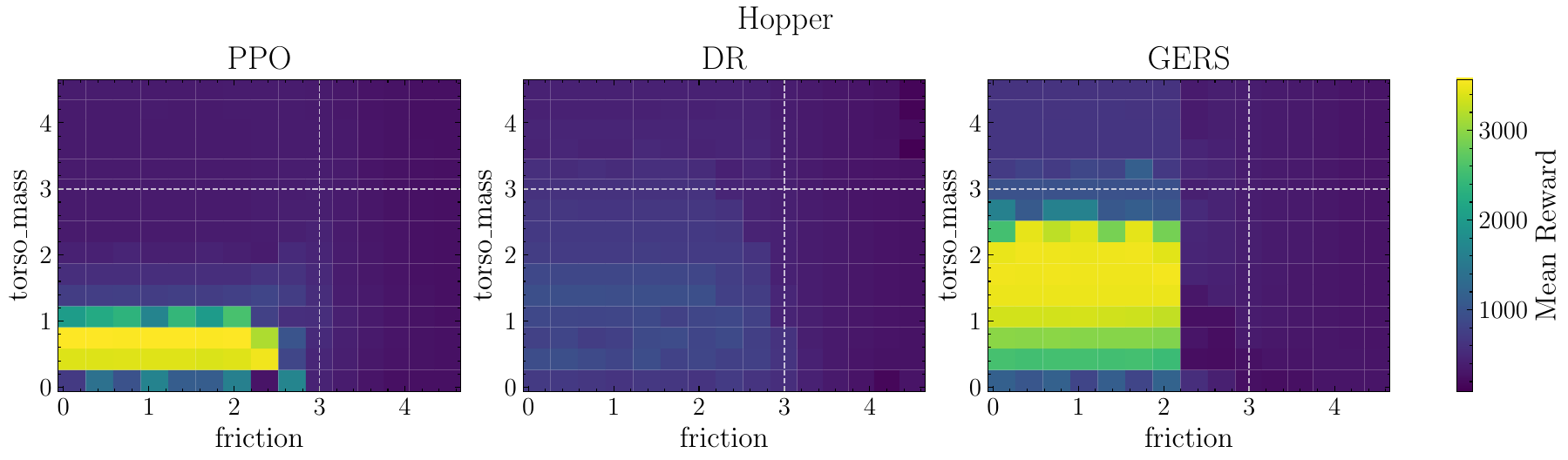}

\vspace{-0.15em}

\includegraphics[width=0.88\textwidth]{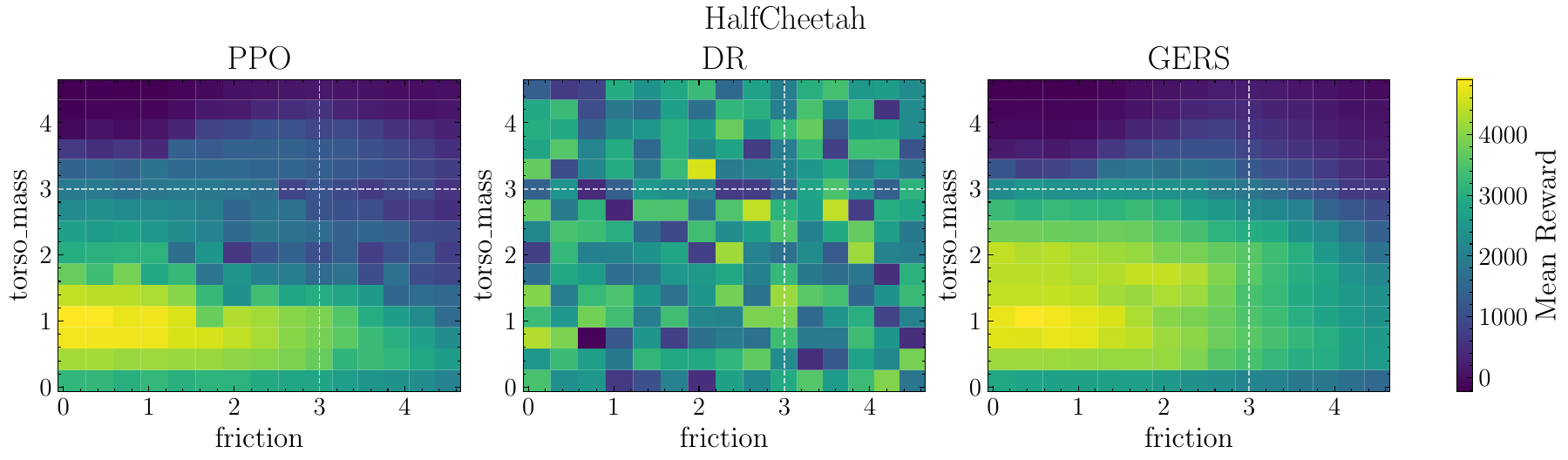}

\vspace{-0.15em}

\includegraphics[width=0.88\textwidth]{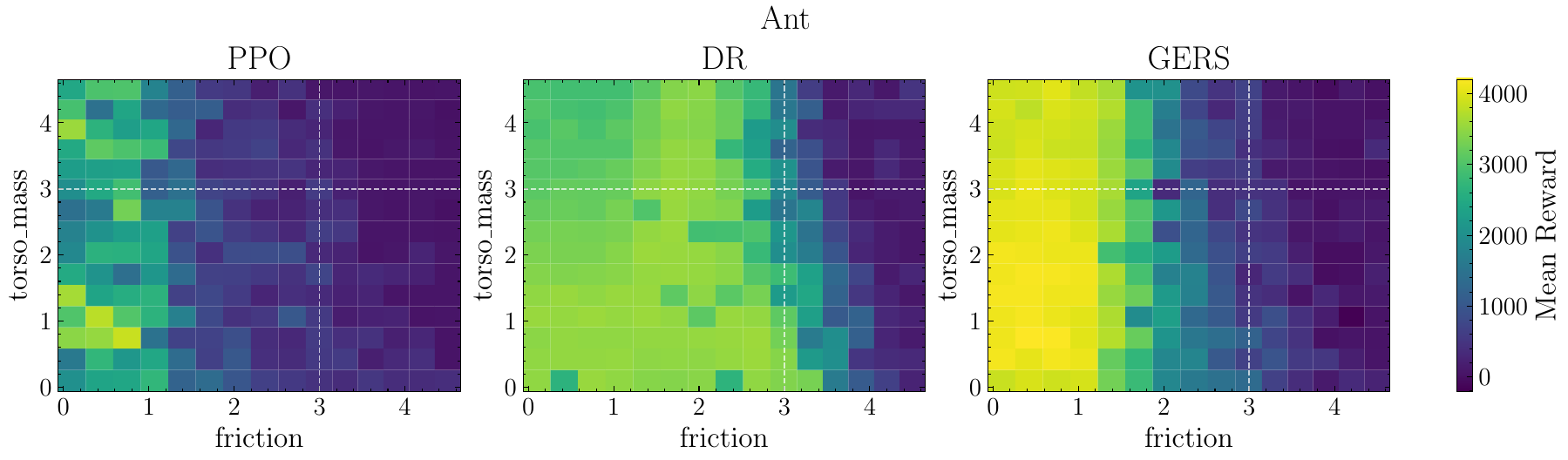}

\caption{Heatmap comparison of PPO, domain randomization (DR), and GERS across MuJoCo tasks under variations of physical parameters. Each heatmap reports the mean test return over the parameter grid. The white dashed lines demarcate the boundary between the in-distribution (in-dist) and out-of-distribution (OOD) evaluation regions.}
\label{fig:heatmap_results}

\end{figure*}

To thoroughly test policy robustness, we evaluate each policy across a wide range of physical conditions on a $15 \times 15$ parameter grid (\Cref{fig:heatmap_results}), reporting the mean return per policy. For each grid environment, we scale two critical parameters (such as friction, mass, or gravity) from $0.1$ up to $4.5$ times their default MuJoCo values. Extending the grid beyond the maximum training value of $3.0$ lets us visualize both in-distribution and out-of-distribution (OOD) performance.
The policy's OOD performance depends heavily on the number of parameters shifted. When only a single parameter exceeds the in-distribution bound ($c > 3.0$) while the other remains within range, the proposed method maintains strong robustness and continues to achieve high returns across most tasks. However, performance drops significantly when both parameters are pushed into the extreme OOD region simultaneously. This pattern indicates that while our method successfully adapts to one extreme physical change, it struggles to generalize when forced to handle multiple extreme variations at once.

\section{Conclusion}
We propose GERS, a CMA-ES-based reward shaping framework that automatically optimizes linear reward weights to improve RL generalization from limited training environments. Across four MuJoCo tasks under varying physical parameters, GERS improves generalization over standard PPO and is competitive with DR, despite far more restrictive data access: unlike DR, which requires interaction trajectories during training, GERS targets the setting where such trajectory access is severely limited.

\subsubsection{Limitations and Future Work}
While \ppabbr~demonstrates strong generalization under restricted data access, our study has four main limitations.
First, our evaluation is restricted to dense-reward continuous control tasks, so it remains unclear how much \ppabbr~contributes when the reward signal is weak or delayed. Future work should evaluate the method on sparse-reward problems, where effective reward shaping is expected to matter most.
Second, the high computational cost ($G\lambda K$) constrained us to a small number of trials, limiting the statistical power of our comparisons. More independent trials are desired to establish the reliability of the observed improvements through proper statistical testing.
Third, our results do not fully isolate the source of the performance gain. Because \ppabbr~trains many PPO policies during the evolutionary search and the best validation policy is selected, the evaluation of the effect of the reward shaping itself is not isolated from the effect of selecting the best validation policy among many stochastically trained policies. An ablation study of the proposed approach with a fixed (unmodified) reward function will isolate the gains of reward shaping from the benefits of best-of-many selection.
Fourth, the upper-level CMA-ES does not fully converge, as is reported in \Cref{sec:convergence_issue}. Future work should implement noise-aware CMA-ES variants to suppress the ranking noise that currently impedes convergence.

\begin{credits}
\subsubsection{\ackname} 
This study was partly funded by JSPS KAKENHI 26K02993.

\subsubsection{\discintname}
The authors have no competing interests to declare that are relevant to the
content of this article.
\end{credits}
\clearpage
\bibliographystyle{splncs04}
\bibliography{ppsn2026}

\include{appendix}
\end{document}

%% file: packages.tex
\usepackage{latexsym}
\usepackage{amsmath}
\usepackage{cite}
\usepackage{amsfonts}
\usepackage{mathtools}       
\usepackage{stmaryrd}        
\usepackage{bm}              
\usepackage{subcaption}
\usepackage{makecell}
\usepackage{tabularx}
\usepackage{booktabs}
\usepackage{graphicx}
\usepackage[font=footnotesize]{caption}
\usepackage{multirow} 
\usepackage{makecell}
\usepackage{changes}

\providecommand{\argmax}{\operatornamewithlimits{arg\,max}} 

\providecommand{\R}{\mathbb{R}} 
\providecommand{\E}{\mathbb{E}} 

\providecommand{\st}{\mathrm{s.t.}}

\DeclarePairedDelimiterX{\inner}[2]{\langle}{\rangle}{#1, #2}

\usepackage{xcolor}

\usepackage{cleveref}

\usepackage{minted}

\usepackage{algorithm}
\usepackage{algpseudocode}

\usepackage{siunitx}

\providecommand{\Jtrh}{\hat{J}_{\mathrm{train}}}
\providecommand{\Jvl}{J_{\mathrm{val}}}

\providecommand{\E}{\mathbb{E}}
\providecommand{\M}{\mathbb{M}}  
\algrenewcommand\algorithmicrequire{\textbf{Input:}}
\algrenewcommand\algorithmicensure{\textbf{Output:}}

\providecommand{\II}{\bm{I}}
\providecommand{\mm}{\bm{m}}

\providecommand{\pphi}{\phi}
\providecommand{\ppabbr}{GERS}
\providecommand{\ppname}{Generalization via Evolutionary Reward Shaping}
\providecommand{\ppnamefull}{\ppname\space(\ppabbr)}

%% file: appendix.tex
\clearpage
\appendix

\crefalias{section}{appendix}
\crefalias{subsection}{appendix}
\crefalias{subsubsection}{appendix}
\crefname{appendix}{Appendix}{Appendices}
\Crefname{appendix}{Appendix}{Appendices}

\begin{figure*}[t]
\centering

\begin{subfigure}[t]{0.45\textwidth}
    \centering
    \includegraphics[width=\linewidth]{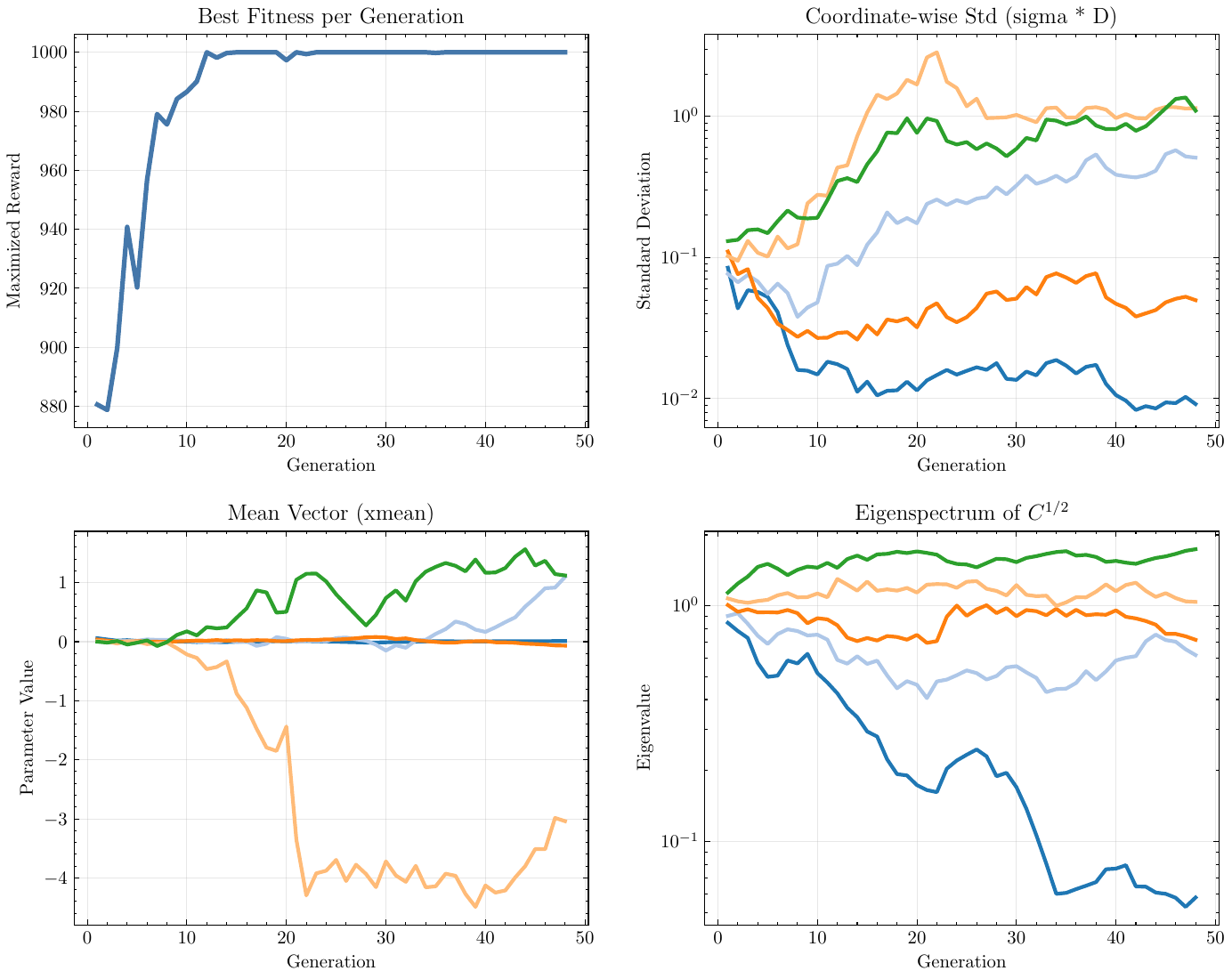}
    \caption{InvertedPendulum}
    \label{fig:cmaes_inv}
\end{subfigure}
\hfill
\begin{subfigure}[t]{0.45\textwidth}
    \centering
    \includegraphics[width=\linewidth]{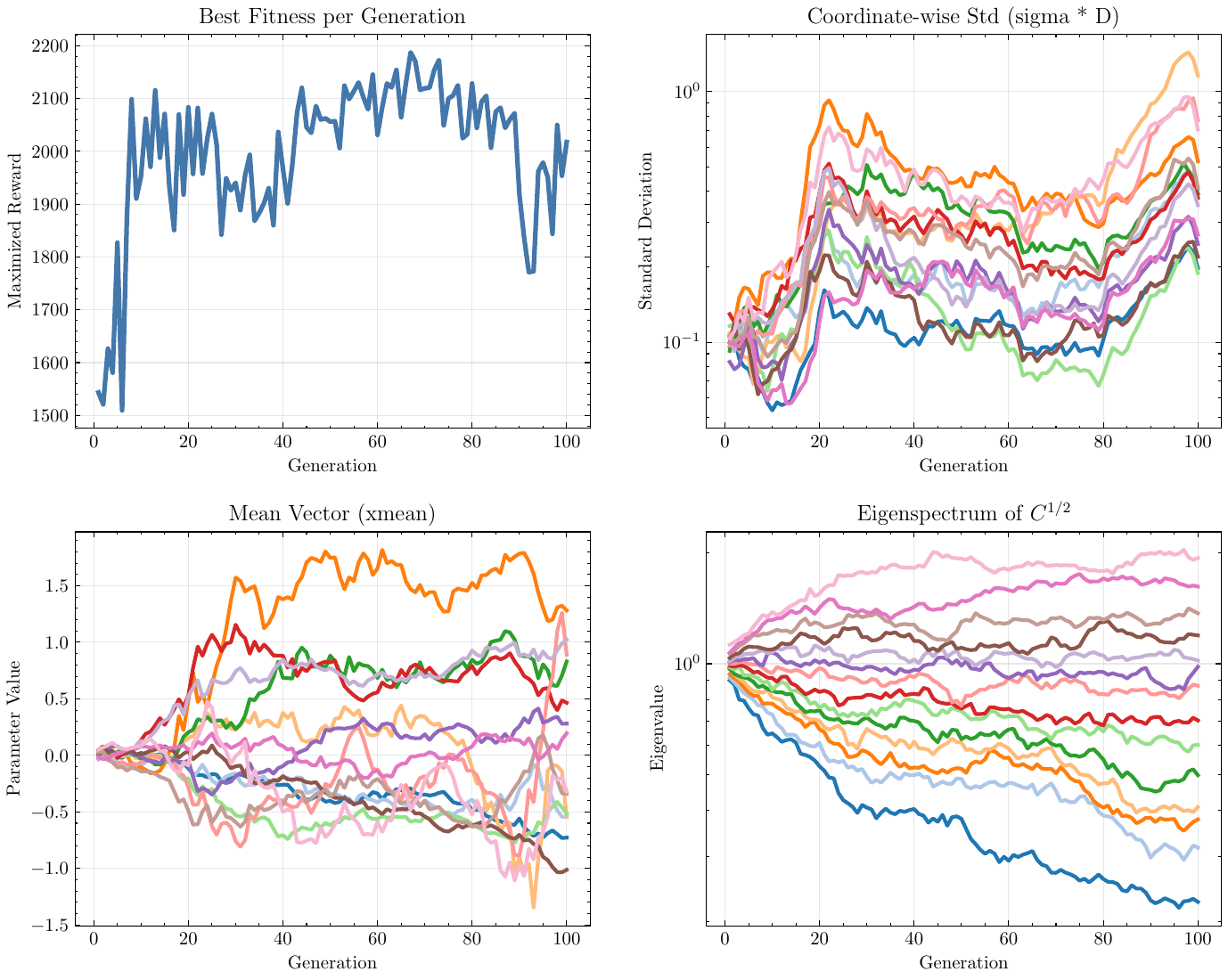}
    \caption{Hopper}
    \label{fig:cmaes_hopper}
\end{subfigure}

\vspace{0.8em}

\begin{subfigure}[t]{0.45\textwidth}
    \centering
    \includegraphics[width=\linewidth]{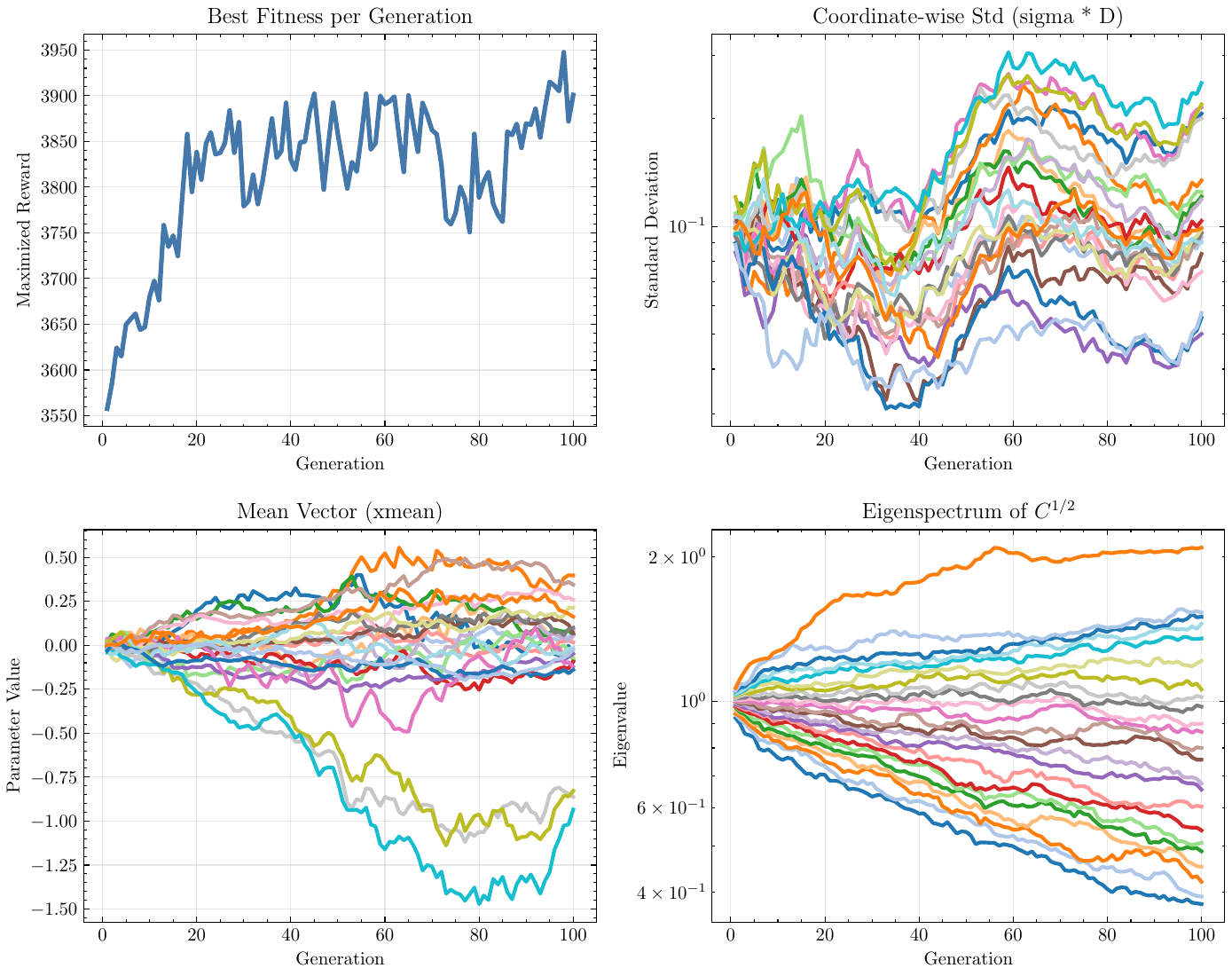}
    \caption{HalfCheetah}
    \label{fig:cmaes_cheetah}
\end{subfigure}
\hfill
\begin{subfigure}[t]{0.45\textwidth}
    \centering
    \includegraphics[width=\linewidth]{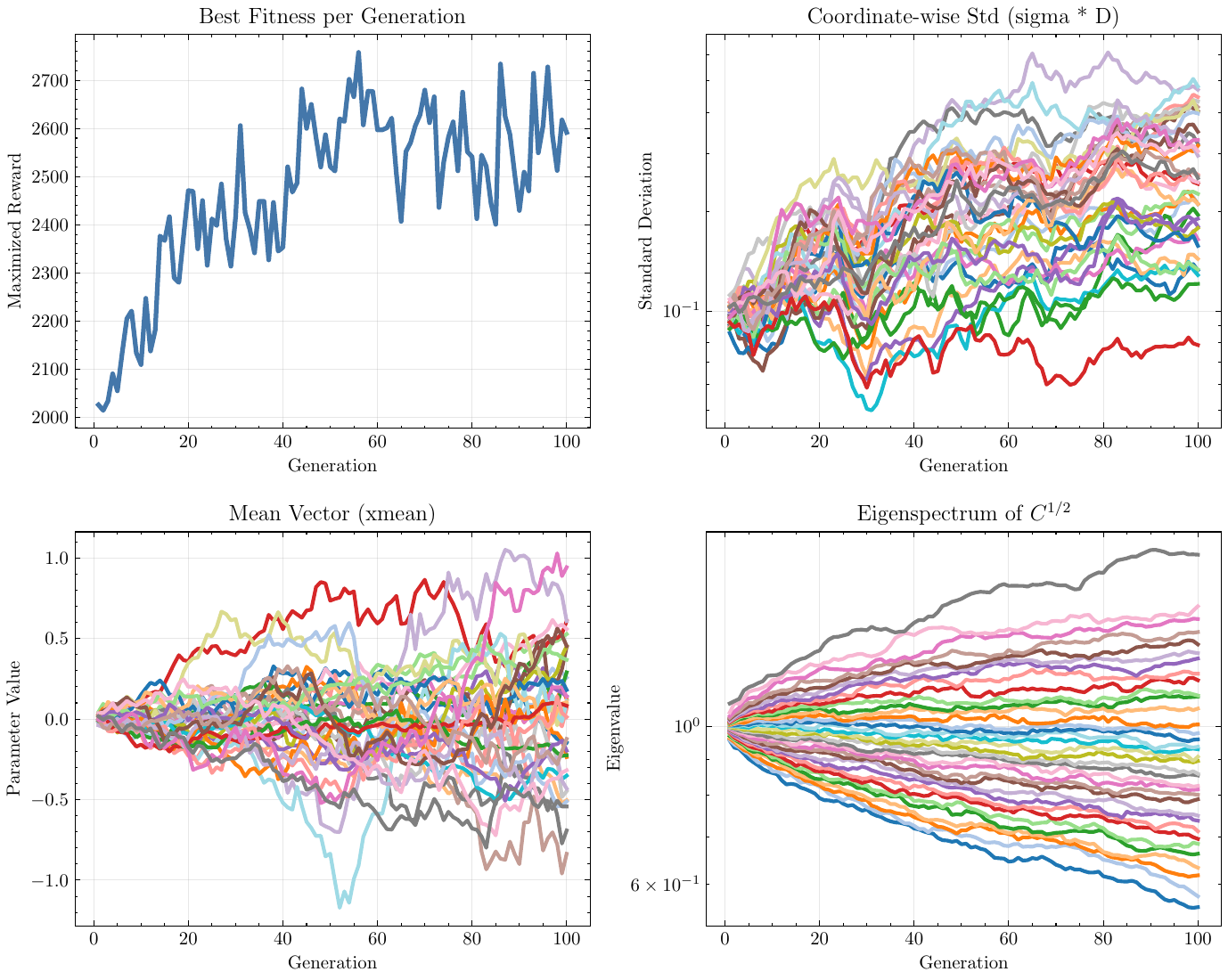}
    \caption{Ant}
    \label{fig:cmaes_ant}
\end{subfigure}

\caption{Bilevel optimization with dd-CMA-ES. Within dd-CMA-ES, the covariance matrix $\Sigma$ of the sampling distribution is decomposed as $\Sigma = \sigma^2 D C D$, where $\sigma$ is the global step size, $C$ is the correlation matrix, and $D$ is a diagonal scaling matrix. When scaled by $\sigma$, the diagonal entries of $D$ represent the coordinate-wise standard deviations of the sampling distribution.}
\label{fig:cmaes_diagnostics}
\end{figure*}

\section{Convergence Issue of CMA-ES}
\label{sec:convergence_issue}
\Cref{fig:cmaes_diagnostics} shows the internal behavior of the CMA-ES reward-shaping process. The validation fitness $f(\nu)$ gradually rises in early generations before plateauing, indicating that CMA-ES finds a good region of reward parameters during training. However, the sampling distribution fails to converge, as indicated by large and non-decreasing coordinate-wise standard deviations, and anisotropy persisting in the covariance matrix. We identify a primary factor for this convergence failure: \textit{insufficient timesteps for lower-level convergence}. Due to computational constraints, estimating the optimal policy parameters $\theta^{\star}(\nu^{(i)})$ by running PPO for only a limited number of timesteps ($K$) typically yields sub-optimal policies. Evaluating these partially trained policies produces inaccurate fitness scores, which in turn generates a false ranking of the candidate reward parameters. Because CMA-ES relies purely on this rank order to update its search distribution, these false rankings act as severe evaluation noise. CMA-ES is highly sensitive to such ranking noise, which misguides the covariance matrix adaptation and impedes upper-level convergence. However, resolving this by drastically increasing $K$ is undesirable, as it introduces the competing risk of overfitting the policy to the training environments and worsening generalization.

\section{Train vs Validation vs Test Behavior}
\begin{figure*}[h]
\centering

\makebox[0.47\textwidth]{\centering InvertedPendulum}
\hspace{0.04\textwidth}
\makebox[0.47\textwidth]{\centering Hopper}

\vspace{0.4em}

\includegraphics[width=0.47\textwidth]{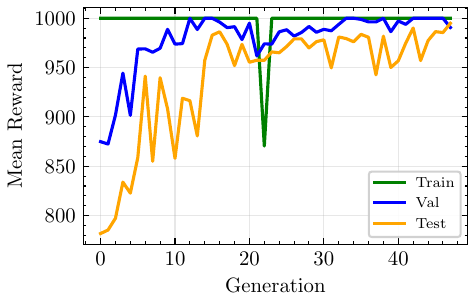}
\hspace{0.04\textwidth}
\includegraphics[width=0.47\textwidth]{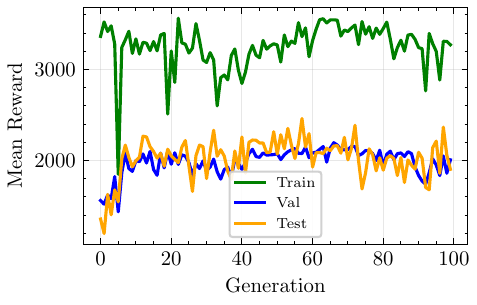}

\vspace{1em}

\makebox[0.47\textwidth]{\centering HalfCheetah}
\hspace{0.04\textwidth}
\makebox[0.47\textwidth]{\centering Ant}

\vspace{0.4em}

\includegraphics[width=0.47\textwidth]{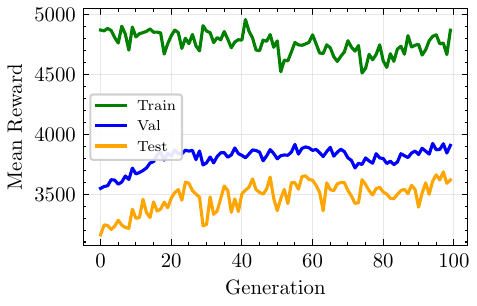}
\hspace{0.04\textwidth}
\includegraphics[width=0.47\textwidth]{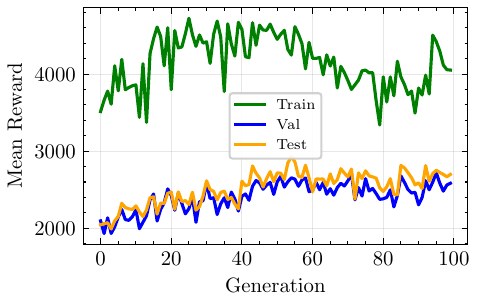}

\caption{\ppabbr~Train vs.\ Validation vs.\ Test Score.}
\label{fig:gers_train_val_test}

\end{figure*}

We observe that the improvement on the test environments closely follows the trend of the CMA-ES validation fitness curve \Cref{fig:gers_train_val_test}. After CMA-ES begins to increase the validation score, the test performance also increases in a similar way. Importantly, there is no sign of divergence between validation and test performance. This consistency indicates that the policy are not overfitting to the validation set, but instead, capture the signals that transfer well to unseen environments by
the shaped reward that helps guide the policy toward behaviors that remain effective across different unseen environments.

\begin{table*}[t]
\centering
\caption{Wilcoxon rank-sum test comparing GERS against PPO and DR using the 5 final trial return for each task.}
\label{tab:final_test_wilcoxon}
\small
\begin{tabular*}{\linewidth}{@{\extracolsep{\fill}} l l l}
\toprule
Task & $p$ vs PPO & $p$ vs DR \\
\midrule
InvertedPendulum & $0.0079$ & $0.5476$ \\
Hopper & $0.0079$ & $0.0159$ \\
HalfCheetah & $0.1508$ & $0.5476$ \\
Ant & $0.0952$ & $1.0000$ \\
\bottomrule
\end{tabular*}
\end{table*}

\section{Statistical Analysis}
\label{sec:stats_analyse}
\Cref{tab:final_test_mean_std,tab:final_test_wilcoxon} summarize the final test performances and statistical significance across five trials. While GERS achieves higher mean returns than BasePPO across all evaluated tasks, Wilcoxon rank-sum tests \cite{Wilcoxon1945} indicate that these improvements are not statistically significant when adjusting for multiple comparisons. Specifically, applying a Bonferroni-correction \cite{bonferroni1936teoria}, significance threshold is $\alpha = 0.05 / 8 = 0.00625$. The lowest $p$-values for GERS versus PPO observed on the InvertedPendulum and Hopper tasks ($p = 0.0079$) still exceed the required alpha level. Similarly, the apparent advantage of GERS over DR on the Hopper task ($p = 0.0159$) is not considered significant under this stricter threshold. Furthermore, while DR yields higher raw mean returns than GERS on HalfCheetah and Ant, comparisons between GERS and DR reveal no significant difference ($p = 0.5476$ and $p = 1.0000$, respectively). Ultimately, after correcting for multiple comparisons, the performance variations between the methods do not reach statistical significance, largely due to the limited number of trials constraining the statistical test results.

\begin{table}[t]
\centering
\caption{Environment parameter variation ranges.}
\label{tab:env_altered_params}
\begin{tabular*}{\linewidth}{@{\extracolsep{\fill}} l l l}
\toprule
Environment & Altered Parameters & Range \\
\midrule
InvertedPendulum & Torso mass, Gravity & $[0.1,\,3.0]$ \\
Hopper & Torso mass, Floor friction & $[0.1,\,3.0]$ \\
HalfCheetah & Torso mass, Floor friction & $[0.1,\,3.0]$ \\
Ant & Torso mass, Floor friction & $[0.1,\,3.0]$ \\
\bottomrule
\label{tab:physical_params}
\end{tabular*}
\end{table}

\section{Continuous Space Normalization} \label{sec:normalization}
The environment features a continuous observation space with potentially unbounded value ranges in each dimension.To ensure numerical stability and facilitate efficient learning, we apply a dimension-wise hybrid normalization strategy to the raw continuous state vector $x \in \mathbb{R}^n$ to yield the normalized state vector $\hat{x} \in \mathbb{R}^n$.

\subsection{Unbounded Continuous State Space}
For unbounded continuous state variables, we apply standard Z-score normalization.The parameters $\mu_i$ and $\sigma_i^2$ represent the empirical mean and variance of the $i$-th dimension, respectively. Unlike standard online normalization where these values are continually updated, our approach utilizes pre-computed running statistics obtained from a pretrained PPO model. To ensure consistent state representations, $\mu_i$ and $\sigma_i^2$ are frozen throughout subsequent execution and generations. To maintain numerical stability in cases where the pre-computed variance is near zero, a small constant $\epsilon = 10^{-8}$ is added to the denominator. The normalized value is computed as:
\begin{equation}
\hat{x}_i = \frac{x_i - \mu_i}{\sqrt{\sigma_i^2 + \epsilon}}
\end{equation}

\subsection{Bounded Continuous Action Space}
For environments parameterized with continuous action spaces, we apply a systematic clipping and normalization procedure. This ensures that the actions generated by the policy or environment mechanics are strictly bounded and linearly scaled to the standardized interval $[-1, 1]$.

Let $a \in \mathbb{R}^{d_a}$ denote the raw action vector. Let $H^a_j$ and $L^a_j$ represent the strict upper and lower bounds of the $j$-th action dimension, respectively.

To prevent numerical instability (division by zero) in degenerate dimensions where the upper and lower bounds are identical, we define an adjusted range denominator $D^a_j$:
\begin{equation}
    D^a_j = \begin{cases} 1.0, & \text{if } H^a_j - L^a_j = 0 \\ H^a_j - L^a_j, & \text{otherwise} \end{cases}
\end{equation}
The normalization procedure first explicitly clips the raw action to respect the environmental bounds, and then applies Min-Max scaling. The fully normalized action $\hat{a}_j \in [-1, 1]$ for the $j$-th dimension is computed as:
\begin{equation}
    \hat{a}_j = 2.0 \left( \frac{\text{clip}(a_j, L^a_j, H^a_j) - L^a_j}{D^a_j} \right) - 1.0
\end{equation}
where $\text{clip}(z, \min, \max)$ bounds the continuous value $z$ strictly within the specified limits.

\begin{table*}[h] 
\centering
\caption{Detailed Actor-Critic Network Architecture and Layer Specifications}
\label{tab:network_arch_full}
\begin{tabularx}{\textwidth}{l X c c c}
\toprule
\textbf{Component} & \textbf{Layer} & \textbf{Input Dim} & \textbf{Output Dim} & \textbf{Activation} \\ 
\midrule
\multirow{3}{*}{Shared Feature Extractor} & Linear Layer 1 & $d_{obs}$ & 64 & ReLU \\
 & Linear Layer 2 & 64 & 64 & ReLU \\
 & Linear Layer 3 (Feature Dim) & 64 & 64 & ReLU \\ 
\midrule
\multirow{3}{*}{Actor Head ($\pi_\theta$)} & Hidden Layer 1 & 64 & 64 & ReLU \\
 & Hidden Layer 2 & 64 & 64 & ReLU \\
 & Output (Action Distribution) & 64 & $d_{act}$ & Linear \\ 
\midrule
\multirow{3}{*}{Critic Head ($V_\phi$)} & Hidden Layer 1 & 64 & 64 & ReLU \\
 & Hidden Layer 2 & 64 & 64 & ReLU \\
 & Output (State Value $V$) & 64 & 1 & Linear \\ 
\bottomrule
\end{tabularx}
\end{table*}

\begin{table}[h]
\centering
\caption{Hyperparameters used for CMA-ES reward optimization.}
\label{tab:gers_hyperparameters}
\begin{tabularx}{0.75\linewidth}{X l}
\toprule
Hyperparameter & Value \\
\midrule
Generation & 100 \\

$m$ (CMA-ES initial mean) & 0 \\
$\sigma$ (CMA-ES initial step size) & 0.1 \\
$\lambda$ (CMA-ES population size) & 60 \\
Lower bound & $-5.0$ \\
Upper bound & $5.0$ \\
\bottomrule
\end{tabularx}
\end{table}

\begin{table}[h]
\centering
\caption{PPO hyperparameters used for policy optimization.}
\label{tab:ppo_hyperparameters}

\begin{tabularx}{0.75\linewidth}{X l}
\toprule
Hyperparameter & Value \\
\midrule
Timestep & 81,920 \\
Learning rate & $1\times10^{-4}$ \\
Number of steps ($n_{\text{steps}}$) & 2048 \\
Batch size & 256 \\
Number of epochs ($n_{\text{epochs}}$) & 10 \\
Discount factor ($\gamma$) & 0.99 \\
GAE parameter ($\lambda$) & 0.95 \\
Clipping range & 0.20 \\
Entropy coefficient & 0.00 \\
Value function coefficient & 0.50 \\
Maximum gradient norm & 0.50 \\
\bottomrule
\end{tabularx}

\end{table}

\section{Reproducibility Details}
To ensure the complete reproducibility of our experiments, we detail the specific network architectures and algorithmic hyperparameters utilized across all components of our framework, which are summarized in \Cref{tab:network_arch_full,tab:gers_hyperparameters,tab:ppo_hyperparameters}